\documentclass[conference]{IEEEtran}
\IEEEoverridecommandlockouts
\usepackage{cite}
\usepackage{amsmath,amssymb,amsfonts}
\usepackage{graphicx}
\usepackage{xcolor}

\usepackage{adjustbox}
\usepackage{tabularray}
\usepackage{url}
\usepackage{multirow}
\def\BibTeX{{\rm B\kern-.05em{\sc i\kern-.025em b}\kern-.08em
    T\kern-.1667em\lower.7ex\hbox{E}\kern-.125emX}}

\makeatletter
\newcommand{\newlineauthors}{%
  \end{@IEEEauthorhalign}\hfill\mbox{}\par
  \mbox{}\hfill\begin{@IEEEauthorhalign}
}
\makeatother

\title{CPT-Boosted Wav2vec2.0: Towards Noise Robust Speech Recognition for Classroom Environments}

\author{\IEEEauthorblockN{Ahmed Adel Attia$^1$, Dorottya Demszky$^2$,  Tol\'{u}l\d{o}p\d{\'{e}} \`{O}g\'{u}nr\d{\`{e}}m\'{i}$^2$, Jing Liu$^1$,  Carol Espy-Wilson$^1$\thanks{This work is supported by the Grand Challenge Award at the University of Maryland and through the generous support of the Bill and Melinda Gates Foundation}}
\IEEEauthorblockA{
\textit{$^1$University of Maryland College Park, MD, USA},
\textit{$^2$Stanford University, CA, USA}\\
aadel@umd.edu, ddemszky@stanford.edu, tolulope@cs.stanford.edu,
jliu28@umd.edu,
espy@umd.edu\vspace{-2em}}

}
\begin{document}
\maketitle
\begin{abstract}
Creating Automatic Speech Recognition (ASR) systems that are robust and resilient to classroom conditions is paramount to the development of AI tools to aid teachers and students. In this work, we study the efficacy of continued pretraining (CPT) in adapting Wav2vec2.0 to the classroom domain. We show that CPT is a powerful tool in that regard and reduces the Word Error Rate (WER) of Wav2vec2.0-based models by upwards of 10\%. More specifically, CPT improves the model's robustness to different noises, microphones and classroom conditions.

\end{abstract}

\begin{IEEEkeywords}
self-supervised learning, wav2vec2, asr, automatic speech recognition, classrooms
\end{IEEEkeywords}
\vspace{-5pt}
\section{Introduction}
Developing interactive AI tools for classrooms can help create a more fair and equitable learning environment while providing tools for teachers to aid in the teaching process. Automatic Speech Recognition (ASR) is a critical part of such pipelines. Transcripts generated by ASR systems can be analyzed on many levels to understand the dynamics in the classrooms --- if they are sufficiently accurate \cite{demszky2023improving,jacobs2022promoting,jacobs2024automated}. However, classroom ASR remains a largely unsolved challenge. ASR systems face significant challenges when dealing with children's speech, even under ideal conditions. These systems are predominantly trained on adult speech, which leaves them ill-equipped to handle the distinct characteristics of children's speech. Children generally articulate less clearly than adults \cite{lee1999acoustics}, and their speech possesses unique acoustic and linguistic properties that differ significantly from those of adults \cite{gerosa2009review}. 

In our previous work \cite{attia2023kid}, we analyzed Whisper's\cite{radford2023robust} performance on children's speech and our findings suggest that while it can achieve adult-level performance with simple and scripted prompts—demonstrating its ability to adapt to the acoustic characteristics of children's speech—it still struggles with the linguistic characteristics of children’s speech. Several other studies have shown that finetuning improves the performance of popular ASR systems, yet a significant gap remains between the effectiveness of these systems on adult versus children's speech \cite{attia2023kid, shahnawazuddin2024developing, jain2023adaptation, southwell2024automatic}.

In classrooms, the problem gets more complicated. Babble noise is considered one of the most challenging noises even in adult speech with adult babble \cite{simic2024self}. However, children’s babble noise like in classrooms is more challenging as it is less likely to occur in public datasets that most current ASR models are trained on. Other conditions that affect the accuracy of ASR systems, like far-field speech \cite{zhu2024multichannel}, and multi-speaker conditions \cite{chang2019end, chang2020end} are also abundant in classrooms. 

The previously mentioned challenges are further exacerbated by the scarcity of transcribed classroom datasets, many of which are not publicly available due to the sensitive nature of data involving minors. While non-transcribed classroom recordings exist, transcription costs can be prohibitively high. As such, this low-resource environment is particularly well-suited for self-supervised speech representation models like Wav2vec2.0 \cite{baevski2020wav2vec} and HuBERT \cite{hsu2021hubert}, which can leverage untranscribed data for pretraining, while the available transcribed data can be used for finetuning the model for ASR tasks.

In this paper, we propose continued pretraining (CPT) as an effective way to adapt Wav2vec2.0 to domain-specific noisy speech recognition, namely classroom speech recognition. Starting from different Wav2vec2.0 versions as initialization, we perform self-supervised pretraining on unlabeled noisy classroom data, then finetune on small labeled subsets. 

\textbf{Our contributions} in this research are as follows:
\begin{itemize}
    \item We show that CPT is the most effective tool to adapt Wav2vec2.0 to noisy conditions like classrooms compared to existing methods. 
    \item We perform an ablation study to determine the optimal pretrained model to use as initialization for CPT.
    \item We show how CPT can improve the robustness of Wav2vec2.0, not only to noise but to different microphone configurations and demographics.
    \item We show that our proposed method is more robust to noise than State Of The Art (SOTA) ASR models.
    \item We demonstrate the use of existing classroom text corpora for Language Model (LM) training.
\end{itemize}

To facilitate further research and reproducibility, our training code can be found on the first author's GitHub \footnote{\url{https://github.com/ahmedadelattia/finetune_w2v_fairseq}}.
\vspace{-5pt}
\section{Background and related previous works}

\subsection{Wav2vec2.0}

Wav2vec2.0 is a Self Supervised Learning (SSL) speech representation model developed by \cite{baevski2020wav2vec} which utilizes the contextualization capabilities of transformers to learn contextual self-supervised representations from unlabeled audio. 

While supervised speech models, like Whisper, learn directly on human-annotated, task-specific labeled data to achieve SOTA performance, Wav2vec2.0 models are first \textit{pretrained} on unlabeled data using contrastive learning to extract contextual representation. Then, a single classification layer is added on top of the model and the entire model is \textit{finetuned} on a smaller labeled dataset using CTC loss \cite{graves2006connectionist}. \textit{Continued pretraining (CPT)} refers to performing additional self-supervised pre-training on a model that was already pre-trained before finetuning.

\subsection{Adaptation of Wav2vec2.0 to low-resource languages}
Several research papers  \cite{san2024predicting,paraskevopoulos2023sample, nowakowski2023adapting} investigated the effectiveness of CPT in adapting multi-lingual self-supervised ASR systems like XLSR53 \cite{conneau2020unsupervised} and XLS-R \cite{babu2021xls} to low-resource languages.  The research by \cite{nowakowski2023adapting} developed an ASR system for Ainu, a critically endangered and low-resource language. Starting from XLSR53 which was already pretrained on 56K hours from 53 languages, they performed CPT on 234 hours of Ainu recordings. They describe CPT as ``clearly the most effective way to adapt a speech representation model for a new language". CPT decreased their WER by up to 40\% relative to the unadapted model. 


\vspace{-5pt}
\section{Datasets}
\subsubsection{NCTE}
The NCTE dataset consists of video and audio recordings of 2128 4th and 5th-grade elementary math classrooms collected as part of the National Center for Teacher Effectiveness (NCTE) Main Study \cite{kane2022}. The observations took place between 2010 and 2013 across four districts serving historically marginalized students.

For each classroom, 2 to 3 video and audio recordings exist, from different angles and microphones, each lasting 45 minutes to an hour. The total duration of the recordings from all microphones and classrooms is 5235 hours. We resampled the audio from 44.1KHz to 16KHz and cut it into 20-second chunks. About 10\% of the data was reserved for validation.

Out of these recordings, 6 were randomly chosen to be transcribed to create a low-resource unbalanced problem. Classrooms within this subset have varying demographic makeups. One of these recordings, which is denoted in Table \ref{tab:detailed} as 2619, comes from a far field microphone, to test generalization to unseen microphone configurations. The duration of this subset is about 5.15 hours, with the duration of each file between 45-60 minutes, and it was used for supervised \textit{finetuning}. 

\subsubsection{MPT dataset}
We recorded six classrooms as part of the larger M-Powering Teachers (MPT) dataset we are compiling. Two 8th-grade classrooms from a Washington, DC charter school, labeled \textbf{DC-1} and \textbf{DC-2}, served predominantly African-American and Hispanic low-income students, including special education and English language learners. Two 5th-grade classrooms in Eastlake, Ohio, referred to as \textbf{OH-1} and \textbf{OH-2}, had mostly White students, with some in special education. Lastly, two 6th-grade classrooms from a private school in San Jose, California, labeled \textbf{CA-1} and \textbf{CA-2}, had predominantly White and Asian high-income students, with no special education or English language learners.  In each classroom, five microphones were placed at different places and the audio streams were averaged. The positioning of these microphones varied from one classroom to the other depending on the layout. This resulted in a good capture of all the audio in each classroom but also extremely noisy audio in one particularly noisy classroom, CA-1. We keep this configuration to test the model's ability to handle extremely noisy conditions. We use this dataset to test the effect of CPT on improving performance in different but related domains than the one seen during CPT.
\subsubsection{NCTE-Text}
\label{subsubsec:ncte-text}
NCTE-Text is a dataset of anonymized transcriptions of 1660 classrooms from the NCTE dataset \cite{demszky2022ncte}.  To protect subjects' privacy, all names in the text corpus were de-identified and replaced by their initials. These transcripts were not intended for ASR training and are not suitable for it as they are not verbatim. However, this text corpus can be used to train task-specific lightweight n-gram LM for beam-search decoding of Wav2vec2.0 outputs. To make this data suitable for LM training, we replaced the initials with randomly chosen names. To ensure that the names are unbiased towards race or gender, we referenced a list of the most popular baby names by race between 2011 and 2019 in New York City\footnote{\url{https://data.cityofnewyork.us/Health/Popular-Baby-Names/25th-nujf/data}}. For each classroom transcription, de-identified initials were replaced by names sampled from this list, to ensure equitable representation across gender and racial identity. 
\vspace{-5pt}
\section{Experiments}
\vspace{-5pt}
\subsection{CPT Experiments}
We performed three CPT experiments to contrast the effect of the starting checkpoint for CPT on domain adaptation. We considered three 300M parameter checkpoints of Wav2vec2.0. 
\begin{itemize}
    \item \textbf{W2V-LV60}, pretrained on 60K hours of LibriVox\cite{kearns2014librivox}.
    \item \textbf{W2V-Robust}, pretrained on 60K hours of LibriVox, noisy telephone speech and crowd sourced data.\cite{hsu2021robust}.
    \item \textbf{XLS-R}, pretrained on 436K hours of speech from 128 languages, including English\cite{babu2021xls}.
\end{itemize} 

Finetuning the resulting models serves as an ablation study to determine the best criteria for the starting checkpoint of CPT.  The contrast between the performance of W2V-LV60 and W2V-Robust will showcase the effect of additional out-of-domain (OOD) noisy data during initial pretraining on the efficacy of CPT domain adaptation. The performance of XLS-R will showcase the impact of having more pretraining data from completely different domains and languages. 

We also pretrain Wav2vec2.0 from scratch, which we denote as \textbf{W2V-SCR} to contrast the effect of CPT versus pre-training from scratch. For both pre-training from scratch as well as CPT, we use 5235 hours of untranscribed NCTE audio. We used the official configuration file for pre-training Wav2vec from the Fairseq GitHub repository \footnote{\url{https://github.com/facebookresearch/fairseq/blob/main/examples/wav2vec/}}.

We perform two separate 6-fold cross-validation finetuning on each of the two datasets used for training, leaving one classroom recording for validation in each fold. We finetune each off-the-shelf Wav2vec2.0 model (W2V-LV60, W2V-Robust, and XLS-R), and their CPT counterparts as well as W2V-SCR. We used the configuration file designed for finetuning on 10 hours of Librispeech from the Fairseq GitHub repository. 

We preprocess our data by cutting the audio into chunks of 30 seconds or less depending on the timestamps of the transcription. We normalize the text, removing casing and punctuation as described in the seminal Whisper paper.
\vspace{-5pt}
\subsection{N-Gram LM Training}
We train a 5-gram LM on the deanonymized NCTE-Text dataset. We normalize the training data using the Whisper Normalizer library to match the transcription text.

\section{Results and Discussion}
\begin{table}[t]
\centering
\caption{Average cross-validation results from finetuning various Wav2vec2.0 checkpoints, with and without continued pretraining, compared to the Whisper small English-only checkpoint. ``Whisper-FT" indicates finetuning on the target dataset. WER standard deviations are in brackets. }

\begin{adjustbox}{width=\columnwidth,center}\begin{tabular}{cccc}
\hline
                       &                      & \multicolumn{2}{c}{\textbf{WER(STD)}}                                                \\ \cline{3-4}
\rule{0pt}{2ex}\multirow{-2}{*}{\textbf{Model}} &\multirow{-2}{*}{\textbf{LM}} & \textbf{NCTE}                        & \textbf{MPT}                    \\ \hline
\multicolumn{4}{c}{\rule{0pt}{3ex}\textbf{\textit{Pretraining from Scratch}}} \\
     
\rule{0pt}{2ex}\multirow{2}{*}{\textbf{W2V-SCR}}   & None          &    47.34(5.73)    &  51.39 (6.83)                                     \\
 &  5-gram LM &      30.25(15.44)      &           38.59(12.93)            \\                                              \hline \hline
                 
\multicolumn{4}{c}{\rule{0pt}{3ex}\textbf{\textit{No Continued Pretraining}}} \\
\rule{0pt}{2ex}\multirow{2}{*}{\textbf{W2V-LV60K}}     & None & 39.11(13.01)       & 37.82(12.30)                         \\
      & 5-gram LM  & 30.39(14.48)       & 33.56(10.86)                         \\ \hline

\rule{0pt}{2ex}\multirow{2}{*}{\textbf{XLS-R}} & None          & 38.19(10.39)       & 39.12(13.60)                         \\
                                & 5-gram LM  & 29.02(10.96)       & 32.49(10.76)                         \\\hline

\rule{0pt}{2ex}\multirow{2}{*}{\textbf{W2V-Robust}}     & None & 35.07(11.85)       & 36.36(11.54)                          \\
     & 5-gram LM  & 27.99(13.28)       &           31.49(9.97)                                      
                       \\ \hline \hline

\multicolumn{4}{c}{\rule{0pt}{3ex}\textbf{\textit{Continued Pretraining}}} \\
    \rule{0pt}{2ex} \multirow{2}{*}{\textbf{W2V-LV60K (CPT)}}    & None &            22.52(4.89)        &             32.26(8.92)                        \\
                              & 5-gram LM  &        18.13(5.50)            &          26.72(7.72)                            \\ \hline
    \rule{0pt}{2ex} \multirow{2}{*}{\textbf{XLS-R (CPT)}}                          & None & 26.53(5.13)        & 32.16(10.78)                         \\
      & 5-gram LM  & 19.37(4.91)        & 26.80(8.00)                          \\ \hline
\rule{0pt}{2ex}\multirow{2}{*}{\textbf{W2V-Robust (CPT)}}    & None & 25.04(5.28)        & 30.97(9.99)                          \\
    & 5-gram LM  & \textbf{17.71(5.06)}                 & \textbf{26.50(8.09)}                \\\hline\hline
\rule{0pt}{2ex}\textbf{Whisper}          & -          & 24.46(12.37) & 30.15(12.99)                         \\
\rule{0pt}{2ex}\textbf{Whisper-FT}       & -          & 19.14(6.77)        & 28.53(14.07) 
                                              \\  \hline\hline
\end{tabular}%
\end{adjustbox}
\label{tab:results_main}
\vspace{-15pt}

\end{table}
\subsection{Pretraining from scratch on target domain data}
Table \ref{tab:results_main} shows the average cross-validation WER across all test folds for every model. Pre-training  Wav2vec2.0 from scratch, we get high WER of 30.25\% and 38.59\% in NCTE and MPT, respectively. The performance gap between the two datasets indicates that the NCTE task is easier, which aligns with the noise levels observed in the datasets. 

The standard deviation in the results is quite high, indicating that each recording has unique characteristics. As a result, the folds perform differently on each one. Although we consider classroom recordings to be a single domain, the type of classroom environment—whether collaborative or instructional—affects the noise level and the ratio of teacher to student speech. All of these factors influence performance.
\vspace{-10pt}
\subsection{Off-the-shelf pre-trained models}

Previous work \cite{hsu2021robust} shows that pre-training on target domain data improves performance compared to pre-training on OOD data. However, our results indicate that pre-training from scratch on target data underperforms off-the-shelf models, including W2V-LV60K which is pre-trained entirely on clean speech. Unlike \cite{hsu2021robust}, where in-domain and out-of-domain pre-training datasets were the same size, the OOD pre-training datasets are at least 12 times larger than the target domain pre-training dataset, highlighting the crucial role of pre-training dataset size in learning high-quality speech embeddings.

Without CPT, finetuning W2V-Robust outperforms both XLS-R and W2V-LV60K. W2V-Robust was pre-trained on the same data as W2V-LV60K, plus additional noisy English speech, showing that adding OOD noisy data improves performance. XLS-R, despite being trained on a much larger, cross-lingual dataset, performs worse than W2V-Robust, but better than W2V-LV60K. This supports previous findings \cite{babu2021xls, hsu2021robust} that larger cross-domain corpora can help. However, smaller, domain-relevant data (noisy adult English) performs better. All models still outperform training from scratch on smaller in-domain data.

To summarize, pre-training on larger OOD datasets generally yields better performance than small in-domain pre-training. The closer the pre-training data aligns with the target domain, the better the results—though more data still proves beneficial regardless.

\vspace{-10pt}
\subsection{CPT on target domain data}
Looking at CPT results in the third section of Table \ref{tab:results_main}, we can see that for NCTE with LM decoding, CPT improves WER by up to 12.26\%. This shows that CPT is a powerful tool for domain adaptation when labeled data is scarce and more unlabeled data is available. For the MPT dataset which is different from the pre-training data, CPT is still effective, yielding an improvement of up to 6.84\%, proving that CPT is effective in domain adaptation in a generalizable fashion. It is also noticeable how the standard deviation of the WER decreases in both datasets. This shows that performing CPT on diverse classroom environments and noise conditions improves the ability of the model to generalize to these conditions. 

\textbf{\textit{In terms of the choice of starting point for CPT}}, the order of performance with CPT does not follow the order observed with off-the-shelf models. W2V-LV60K outperforms XLS-R with CPT, meaning that the performance edge XLS-R had from initially from large cross-lingual pre-training does not carry forward with CPT, and initial pretraining on English-only datasets is always superior.   However, W2V-Robust still provides the best performance, showing that initial pretraining on OOD noisy English data yields better speech representations when adapted to the target domain.

The gap between LM-decoded and non-LM decoded results is much larger when pretraining from scratch, suggesting the model learns acoustics but lacks sufficient linguistic representation, which LM decoding compensates for. This indicates that initial pretraining on clean adult speech, even in other languages, captures useful linguistic features not learned from smaller, noisy in-domain data. Notably, there's a 25\% gap in non-LM decoded WER between W2V-LV60K (CPT) and W2V-SCR, highlighting that CPT benefits from clean OOD speech and further refinement on in-domain data with CPT.
\subsection{Comparison with Whisper}
Finally, we can see that with CPT, W2V-Robust outperforms the Whisper checkpoint of comparable size (Whisper-small.en), even with finetuning in both NCTE and MPT datasets. The nature of self-supervised speech models breaks down the problem into three parts: pretraining/CPT, finetuning, and LM decoding. This gives us more flexibility to utilize unmappable representations like unlabeled audio or text that doesn't correspond well to said audio. Without CPT or LM decoding, Wav2vec2.0 performs much worse than Whisper, with WER being higher by 15-19\% in the NCTE dataset. The flexibility that Wav2vec2.0 allows beyond supervised finetuning improved the performance by up to 19\% through a combination of CPT and LM decoding. 
\subsection{Detailed analysis of cross-validation results}
\begin{table}[t]
\centering

\caption{Detailed results on each fold of the cross-validation on both the NCTE and MPT datasets, for the off-the-shelf and the CPT version of W2V-Robust and the finetuned small English-only Whisper checkpoint. "Fold" refers to the file used for testing.}
\begin{adjustbox}{width=\columnwidth,center}
\begin{tabular}{cccc}
\hline
\multicolumn{4}{c}{\rule{0pt}{2ex}\textbf{NCTE}}                                                      \\
\hline
\rule{0pt}{2ex}\textbf{Fold}    & \textbf{Whisper-FT} & \multicolumn{1}{c}{\textbf{W2V-Robust}} & \textbf{W2V-Robust (CPT)} \\\hline
\rule{0pt}{2ex}\textbf{144}  & \textbf{14.78}      &              19.86                        & 15.20                     \\
\rule{0pt}{2ex}\textbf{622}  & \textbf{16.97}      &              26.31                        & 18.77                     \\
\rule{0pt}{2ex}\textbf{2619} & 28.4                &              54.03                        & \textbf{27.15}            \\
\rule{0pt}{2ex}\textbf{2709} & \textbf{12.74}      &              19.09                        & 13.12                     \\
\rule{0pt}{2ex}\textbf{2944} & 26.98               &              28.07                        & \textbf{17.61}            \\
\rule{0pt}{2ex}\textbf{4724} & 14.95               &              20.55                        & \textbf{14.43}            \\
\hline
\rule{0pt}{2ex}\textbf{Average}       & 19.14               &          27.99                            & \textbf{17.71}            \\
\hline \hline
\multicolumn{4}{c}{\rule{0pt}{2ex}\textbf{MPT}}                                                  \\\hline
\rule{0pt}{2ex}\textbf{Fold}      & \textbf{Whisper-FT} & \multicolumn{1}{c}{\textbf{W2V-Robust}} & \textbf{W2V-Robust (CPT)} \\\hline
\rule{0pt}{2ex}\textbf{OH-1}                          & 19.76               &             18.55                         & \textbf{15.42}            \\
\rule{0pt}{2ex}\textbf{OH-2}                          & \textbf{16.42}      &             19.89                         & 16.88                     \\
\rule{0pt}{2ex}\textbf{DC-1}                          & 28.79               &             29.42                         & \textbf{24.90}            \\
\rule{0pt}{2ex}\textbf{DC-2}                          & \textbf{26.42}      &             37.32                         & 30.34                     \\
\rule{0pt}{2ex}\textbf{CA-1}                          & 55.77               &             49.03                         & \textbf{41.54}            \\
\rule{0pt}{2ex}\textbf{CA-2}                          & \textbf{24.04}      &             34.47                         & 29.91                     \\
\hline
\rule{0pt}{2ex}\textbf{Average}                           & 28.53               &         31.45                             & \textbf{26.50}      
\\\hline\hline
\end{tabular}%
\end{adjustbox}
\label{tab:detailed}
\vspace{-15pt}

\end{table}
\vspace{-5pt}
In this section, discuss the results in Table \ref{tab:detailed} which shows the WER of each fold of the cross-validation in Whisper-FT and W2V-Robust with and without CPT. 
\subsubsection{NCTE} CPT significantly improves the performance in each fold of the cross-validation with one notable example, recording 2619. This recording is the only one that comes from a far-field microphone, with the entirety of the training data in this cross-validation fold coming from near-field microphones. It is thus no surprise that without CPT, the model performs much worse in this fold than the others. However, with CPT, the error is cut in half, with an absolute improvement of almost 27\%, proving that CPT helps the model generalize to microphone configurations unseen in the labeled data but present in the unlabeled pretraining data. 

In terms of comparison with Whisper, looking at the NCTE results, CPT allows the model to achieve close performance or improve upon Whisper in every fold, with one major improvement noticed with recording 2944. Upon manual inspection, it was noted that this class started as an instructional class for 15 minutes and then the teacher assigned a set of questions to the students and started making rounds in the class. This resulted in a much noisier environment than other classrooms with a higher degree of children's babble noise. Whisper was unable to deal with children's babble noise, often interleaving the target speaker with whatever it could discern from the background noise and sometimes exhibiting characteristic Whisper hallucinations by repeating a single word or phrase, for example, \textit{``how do you know that what what is the denominator what is the denominator the the the the the..."} with the word \textit{"the"} repeating 206 times. Wav2vec does not suffer from the same hallucination problem, and with CPT, it's much more capable of dealing with children's babble noise and focusing on the target speaker correctly. 


\subsubsection{MPT dataset}
The differences in WER between Whisper and W2V-Robust with CPT on each fold are higher in the MPT dataset. W2V-Robust with CPT outperforms Whisper in 3 folds, with the main improvement coming from the recording of class CA-1. This recording is perhaps the noisiest of all the datasets used in this study, as discussed in the \textbf{Dataset} section. Both W2V-Robust and Whisper have high WER for this class recording, but even the non-CPT W2V-Robust outperformed Whisper. Upon manual inspection of the transcriptions, we again see that Whisper suffers from extreme hallucinations with high children babble noise. On the other hand, W2V-Robust with CPT is more capable of handling extremely noisy conditions, outperforming Whisper in this fold by 14\%.

\vspace{-5pt}
\section{Conclusions and Future Work}
\vspace{-5pt}
We have demonstrated how CPT is the most effective method to adapt Wav2vec2.0 models to different domains, when compared to existing methods, improving the model's ability to generalize to different noise conditions. We show that when finetuning exclusively on near-field recordings, CPT cuts the WER of far-field recordings in half, showing that the model learns to generalize to acoustic conditions not found in the labeled set. We've shown that CPT can improve the WER on noisy classroom data by up to 12.26\% on average and up to 27\% in specific conditions. Our results suggest that CPT should be the baseline for low-resource domain adaptation experiments, especially in noisy applications as it is superior to other configurations and SOTA models like Whisper. We provide guidelines for selecting a CPT starting point, emphasizing that domain similarity is more important than the amount of pre-training data. We also propose a race-aware deanonymized classroom text dataset for LM training. 

There is an urgent need for more balanced labeled classroom datasets. To that end, we are developing tools to sample recordings from the unlabeled NCTE dataset for transcription in a way that ensures balanced demographics and fair representation. Lastly, we plan to expand on the work of \cite{zhu2022noise}, using speech enhancement-based Wav2vec2.0 pretraining. We are working on simulating classroom noises to augment clean speech to develop educational tools for physical and virtual classrooms.


\begin{thebibliography}{00}


\bibitem{jacobs2024automated} Jacobs, Jennifer, Scornavacco, Karla, Clevenger, Charis, Suresh, Abhijit, Sumner, Tamara. " Automated feedback on discourse moves: teachers’ perceived utility of a professional learning tool," 
 Educational technology research and development  pp. 1--23, 2024.\\
\bibitem{jacobs2022promoting} Jacobs, Jennifer, Scornavacco, Karla, Harty, Charis, Suresh, Abhijit, Lai, Vivian, Sumner, Tamara. "Promoting rich discussions in mathematics classrooms: Using personalized, automated feedback to support reflection and instructional change," Teaching and Teacher Education vol. 112, no. , pp. 103631, 2022.
\\ 
\bibitem{demszky2023improving} Demszky, Dorottya, Liu, Jing, Hill, Heather C., Sanghi, Shyamoli, Chung, Ariel. "Improving Teachers’ Questioning Quality through Automated Feedback: A Mixed-Methods Randomized Controlled Trial in Brick-and-Mortar Classrooms," EdWorkingPapers   2023.
\\ 
\bibitem{attia2023kid} Attia, Ahmed Adel, Liu, Jing, Ai, Wei, Demszky, Dorottya, Espy-Wilson, Carol. "Kid-Whisper: Towards Bridging the Performance Gap in Automatic Speech Recognition for Children VS. Adults," arXiv preprint arXiv:2309.07927   2023.
\\ 
\bibitem{radford2023robust} Radford, Alec, Kim, Jong Wook, Xu, Tao, Brockman, Greg, McLeavey, Christine, Sutskever, Ilya. "Robust speech recognition via large-scale weak supervision,"   pp. 28492--28518, 2023.
\\ 
\bibitem{baevski2020wav2vec} Baevski, Alexei, Zhou, Yuhao, Mohamed, Abdelrahman, Auli, Michael. "wav2vec 2.0: A framework for self-supervised learning of speech representations," Advances in neural information processing systems vol. 33, no. , pp. 12449--12460, 2020.
\\
\bibitem{hsu2021hubert}Hsu, W., Bolte, B., Tsai, Y., Lakhotia, K., Salakhutdinov, R. \& Mohamed, A. Hubert: Self-supervised speech representation learning by masked prediction of hidden units. {\em IEEE/ACM Transactions On Audio, Speech, And Language Processing}. \textbf{29} pp. 3451-3460 (2021)


\bibitem{zhu2022noise} Zhu, Qiu-Shi, Zhang, Jie, Zhang, Zi-Qiang, Wu, Ming-Hui, Fang, Xin, Dai, Li-Rong. "A noise-robust self-supervised pre-training model based speech representation learning for automatic speech recognition,"   pp. 3174--3178, 2022.
\\ 
\bibitem{demszky2022ncte} Demszky, Dorottya, Hill, Heather. "The NCTE transcripts: A dataset of elementary math classroom transcripts," arXiv preprint arXiv:2211.11772   2022.
\\ 
\bibitem{nowakowski2023adapting} Nowakowski, Karol, Ptaszynski, Michal, Murasaki, Kyoko, Nieuwa{\.z}ny, Jagna. "Adapting multilingual speech representation model for a new, underresourced language through multilingual finetuning and continued pretraining," Information Processing \& Management vol. 60, no. 2, pp. 103148, 2023.
\\ 
\bibitem{san2024predicting} San, Nay, Paraskevopoulos, Georgios, Arora, Aryaman, He, Xiluo, Kaur, Prabhjot, Adams, Oliver, Jurafsky, Dan. "Predicting positive transfer for improved low-resource speech recognition using acoustic pseudo-tokens," arXiv preprint arXiv:2402.02302   2024.
\\
\bibitem{paraskevopoulos2023sample}Paraskevopoulos, G., Kouzelis, T., Rouvalis, G., Katsamanis, A., Katsouros, V. \& Potamianos, A. Sample-Efficient Unsupervised Domain Adaptation of Speech Recognition Systems: A Case Study for Modern Greek. {\em IEEE/ACM Transactions On Audio, Speech, And Language Processing}. (2023)
\\ 
\bibitem{zhu2024multichannel} Zhu, Qiushi, Zhang, Jie, Gu, Yu, Hu, Yuchen, Dai, Lirong. "Multichannel AV-wav2vec2: A Framework for Learning Multichannel Multi-Modal Speech Representation," arXiv preprint arXiv:2401.03468   2024.
\\ \bibitem{shahnawazuddin2024developing} Shahnawazuddin, S, others. "Developing children's ASR system under low-resource conditions using end-to-end architecture," Digital Signal Processing vol. 146, no. , pp. 104385, 2024.
\\ 
\bibitem{jain2023adaptation} Jain, Rishabh, Barcovschi, Andrei, Yiwere, Mariam, Corcoran, Peter, Cucu, Horia. "Adaptation of Whisper models to child speech recognition," arXiv preprint arXiv:2307.13008   2023.
\\ 
\bibitem{southwell2024automatic} Southwell, Rosy, Ward, Wayne, Trinh, Viet Anh, Clevenger, Charis, Clevenger, Clay, Watts, Emily, Reitman, Jason, D’Mello, Sidney, Whitehill, Jacob. "Automatic Speech Recognition Tuned for Child Speech in the Classroom,"   pp. 12291--12295, 2024.
\\ 
\bibitem{gerosa2009review} Gerosa, Matteo, Giuliani, Diego, Narayanan, Shrikanth, Potamianos, Alexandros. "A review of ASR technologies for children's speech,"   pp. 1--8, 2009.
\\ 
\bibitem{lee1999acoustics} Lee, Sungbok, Potamianos, Alexandros, Narayanan, Shrikanth. "Acoustics of children’s speech: Developmental changes of temporal and spectral parameters," The Journal of the Acoustical Society of America vol. 105, no. 3, pp. 1455--1468, 1999.
\\ \bibitem{chang2020end} Chang, Xuankai, Zhang, Wangyou, Qian, Yanmin, Le Roux, Jonathan, Watanabe, Shinji. "End-to-end multi-speaker speech recognition with transformer,"   pp. 6134--6138, 2020.
\\ \bibitem{chang2019end} Chang, Xuankai, Qian, Yanmin, Yu, Kai, Watanabe, Shinji. "End-to-end monaural multi-speaker ASR system without pretraining,"   pp. 6256--6260, 2019.
\\ \bibitem{jain2024exploring} Jain, Rishabh, Barcovschi, Andrei, Yiwere, Mariam Yahayah, Corcoran, Peter, Cucu, Horia. "Exploring Native and Non-Native English Child Speech Recognition With Whisper," IEEE Access vol. 12, no. , pp. 41601--41610, 2024.
\\ \bibitem{conneau2020unsupervised} Conneau, Alexis, Baevski, Alexei, Collobert, Ronan, Mohamed, Abdelrahman, Auli, Michael. "Unsupervised cross-lingual representation learning for speech recognition," arXiv preprint arXiv:2006.13979   2020.
\\ 
\bibitem{hsu2021robust} Hsu, Wei-Ning, Sriram, Anuroop, Baevski, Alexei, Likhomanenko, Tatiana, Xu, Qiantong, Pratap, Vineel, Kahn, Jacob, Lee, Ann, Collobert, Ronan, Synnaeve, Gabriel, others. "Robust wav2vec 2.0: Analyzing domain shift in self-supervised pre-training," arXiv preprint arXiv:2104.01027   2021.
\\ 
\bibitem{babu2021xls} Babu, Arun, Wang, Changhan, Tjandra, Andros, Lakhotia, Kushal, Xu, Qiantong, Goyal, Naman, Singh, Kritika, von Platen, Patrick, Saraf, Yatharth, Pino, Juan, others. "XLS-R: Self-supervised cross-lingual speech representation learning at scale," arXiv preprint arXiv:2111.09296   2021.
\\ 
\bibitem{kane2022} Kane, Thomas, Hill, Heather, Staiger, Douglas. "National Center for Teacher Effectiveness Main Study,"    2022.
\\ 
\bibitem{kearns2014librivox} Kearns, Jodi. "Librivox: Free public domain audiobooks," Reference Reviews vol. 28, no. 1, pp. 7--8, 2014.
\\ 
\bibitem{martin2023bias} Martin, Joshua L, Wright, Kelly Elizabeth. "Bias in automatic speech recognition: The case of African American Language," Applied Linguistics vol. 44, no. 4, pp. 613--630, 2023.
\\ 
\bibitem{garofolo1993timit} Garofolo, John S. "Timit acoustic phonetic continuous speech corpus," Linguistic Data Consortium, 1993.
\\

\bibitem{simic2024self} Simic, Christopher, Bocklet, Tobias. "Self-Supervised Adaptive AV Fusion Module for Pre-Trained ASR Models,"   pp. 12787--12791, 2024.
\\
\bibitem{graves2006connectionist}Graves, A., Fernández, S., Gomez, F. \& Schmidhuber, J. Connectionist temporal classification: labelling unsegmented sequence data with recurrent neural networks. {\em Proceedings Of The 23rd International Conference On Machine Learning}. pp. 369-376 (2006)


\end{thebibliography}
\end{document}